\documentclass[twocolumn]{article}

\usepackage{amsmath,amssymb,alex}

\usepackage{algorithmic}
\usepackage{algorithm}
\usepackage{theorem}
\usepackage{amsmath}
\usepackage{amssymb}

\usepackage{cite}
\usepackage{graphicx}
\usepackage{url}

\newtheorem{theorem}{Theorem}[section]
\newtheorem{lemma}[theorem]{Lemma}

\newenvironment{proof}[1][Proof]{\begin{trivlist}
\item[\hskip \labelsep {\bfseries #1}]}{\end{trivlist}}

\newcommand{\qed}{\nobreak \ifvmode \relax \else
      \ifdim\lastskip<1.5em \hskip-\lastskip
      \hskip1.5em plus0em minus0.5em \fi \nobreak
      \vrule height0.75em width0.5em depth0.25em\fi}

\begin{document}
\title{Learning Graph Matching}

\author{Tib\'erio S. Caetano,
Julian J. McAuley,
Li Cheng,
Quoc V. Le and
Alex J. Smola\thanks{Tib\'erio Caetano, Julian McAuley, Li Cheng, and Alex Smola are with the Statistical Machine Learning Program at NICTA, and the Research School of Information Sciences and Engineering, Australian National University. Quoc Le is with the Department of Computer Science, Stanford University.}}

\maketitle

\begin{abstract}

As a fundamental problem in pattern recognition, graph matching has applications in a variety of fields, from computer vision to computational biology. In graph matching, patterns are modeled as graphs and pattern recognition amounts to finding a correspondence between the nodes of different graphs. Many formulations of this problem can be cast in general as a quadratic assignment problem, where a linear term in the objective function encodes node compatibility and a quadratic term encodes edge compatibility. The main research focus in this theme is about designing efficient algorithms for approximately solving the quadratic assignment problem, since it is NP-hard. In this paper we turn our attention to a different question: how to \emph{estimate} compatibility functions such that the solution of the resulting graph matching problem best matches the expected solution that a human would manually provide. We present a method for \emph{learning graph matching}: the training examples are pairs of graphs and the `labels' are matches between them. Our experimental results reveal that learning can substantially improve the performance of standard graph matching algorithms. In particular, we find that simple linear assignment with such a learning scheme outperforms Graduated Assignment with bistochastic normalisation, a state-of-the-art quadratic assignment relaxation algorithm.

\end{abstract}

\section{Introduction}

Graphs are commonly used as abstract representations for complex structures, including DNA sequences, documents, text, and images. In particular they are extensively used in the field of computer vision, where many problems can be formulated as an attributed graph matching problem. Here the nodes of the graphs correspond to local features of the image and edges correspond to relational aspects between features (both nodes and edges can be attributed, i.e.~they can encode feature vectors). Graph matching then consists of finding a correspondence between nodes of the two graphs such that they 'look most similar' when the vertices are labeled according to such a correspondence.

Typically, the problem is mathematically formulated as a quadratic assignment problem, which consists of finding the assignment that maximizes an objective function encoding local compatibilities (a linear term) and structural compatibilities (a quadratic term). The main body of research in graph matching has then been focused on devising more accurate and/or faster algorithms to solve the problem approximately (since it is NP-hard); the compatibility functions used in graph matching are typically handcrafted.

An interesting question arises in this context: If we are given two attributed graphs to match, $G$ and $G'$, should the optimal match be uniquely determined? For example, assume first that $G$ and $G'$ come from two images acquired by a surveillance camera in an airport's lounge; now, assume the same $G$ and $G'$ instead come from two images in a photographer's image database; should the optimal match be the same in both situations? If the algorithm takes into account exclusively the graphs to be matched, the optimal solutions will be the same\footnote{Assuming there is a single optimal solution and that the algorithm finds it.} since the graph pair is the same in both cases. This is the standard way graph matching is approached today.

In this paper we address what we believe to be a limitation of this approach. We argue that if we know the `conditions' under which a pair of graphs has been extracted, then we should take into account \emph{how graphs arising in those conditions are typically matched}. However, we do not take the information on the conditions explicitly into account, since this would obviously be impractical. Instead, we approach the problem purely from a statistical inference perspective. First, we extract graphs from a number of images acquired under the same conditions as those for which we want to solve, whatever the word `conditions' means (e.g.~from the surveillance camera or the photographer's database). We then \emph{manually} provide what we understand to be the optimal matches between the resulting graphs. This information is then used in a \emph{learning} algorithm which learns a map from the space of pairs of graphs to the space of matches.

In terms of the quadratic assignment problem, this learning algorithm amounts to (in loose language) adjusting the node and edge compatibility functions such that the expected optimal match in a test pair of graphs agrees with the expected match they would have had, had they been in the training set. In this formulation, the learning problem consists of a convex, quadratic program which is readily solvable by means of a column generation procedure.

We provide experimental evidence that applying learning to standard graph matching algorithms significantly improves their performance. In fact, we show that learning improves upon non-learning results so dramatically that linear assignment \emph{with learning} outperforms Graduated Assignment with bistochastic normalisation, a state-of-the-art quadratic assignment relaxation algorithm. Also, by introducing learning in Graduated Assignment itself, we obtain results that improve both in accuracy and speed over the best existing quadratic assignment relaxations. 


A preliminary version of this paper appeared in \cite{CaeCheLeSmo07}.

\subsection{Related Literature}
\label{sec:literature}

The graph matching literature is extensive, and many different types of approaches have been proposed, which mainly focus on approximations and heuristics for the quadratic assignment problem. An incomplete list includes 
\emph{spectral} methods \cite{LeoHeb05,Wang04,Shapiro92,Carcassoni03,Caelli04}, \emph{relaxation labeling} and \emph{probabilistic approaches}
\cite{Caetano04Graphical,Rosenfeld82,Wilson97,Hancock02,Christmas95,Kittler89,Li94}, \emph{semidefinite relaxations} \cite{Schellewald04}, \emph{replicator equations} \cite{Pelillo99a}, \emph{tree search} \cite{Messmer98}, and \emph{graduated assignment} \cite{Gold96}. Spectral methods consist of studying the similarities between the spectra of the adjacency or Laplacian matrices of the graphs and using them for matching. Relaxation and probabilistic methods define a probability distribution over mappings, and optimize using discrete relaxation algorithms or variants of belief propagation. Semidefinite relaxations solve a convex relaxation of the original combinatorial problem. Replicator equations draw an analogy with models from biology where an equilibrium state is sought which solves a system of differential equations on the nodes of the graphs. Tree-search techniques in general have worst case exponential complexity and work via sequential tests of compatibility of local parts of the graphs. Graduated Assignment combines the `softassign' method \cite{Rangarajan99} with Sinkhorn's method \cite{Sinkhorn64} and essentially consists of a series of first-order approximations to the quadratic assignment objective function. This method is particularly popular in computer vision since it produces accurate results while scaling reasonably in the size of the graph.


The above literature strictly focuses on trying better \emph{algorithms} for approximating a solution for the graph matching problem, but does not address the issue of how to determine the compatibility functions in a principled way.

In \cite{Pelillo94} the authors learn compatibility functions for the relaxation labeling process; this is however a different problem than graph matching, and the `compatibility functions' have a different meaning. Nevertheless it does provide an initial motivation for learning in the context of matching tasks. In terms of methodology, the paper most closely related to ours is possibly \cite{LacTasKleJor06}, which uses structured estimation tools in a quadratic assignment setting for word alignment. A recent paper of interest shows that \emph{very} significant improvements on the performance of graph matching can be obtained by an appropriate \emph{normalization} of the compatibility functions \cite{CouSriShi07}; however, no learning is involved.

\begin{table}
  \hrule 
  \smallskip
  \caption{Definitions and Notation}
  \label{tab:definition}
  \smallskip
  \hrule 
  \smallskip
 $G$ - generic graph (similarly, $G'$);\\
 $G_{i}$ - attribute of node $i$ in $G$ (similarly, $G'_{i'}$ for $G'$);\\
 $G_{ij}$ - attribute of edge $ij$ in $G$ (similarly, $G'_{i'j'}$ for $G'$);\\
 $\mathcal G$ - space of graphs ($\mathcal G \times \mathcal G$ - space of pairs of graphs);\\
 $x$ - generic observation: graph pair $(G,G')$; $x\in\Xcal$, space of observations;\\
 $y$ - generic label: matching matrix; $y\in\Ycal$, space of labels;\\
 $n$ - index for training instance; $N$ - number of training instances;\\
 $x^n$ - $n^{th}$ training observation: graph pair $(G^n,G'^n)$;\\
 $y^n$ - $n^{th}$ training label: matching matrix;\\
 $g$ - predictor function;\\
 $y^w$ - optimal prediction for $g$ under $w$;\\
 $f$ - discriminant function;\\
 $\Delta$ - loss function;\\
 $\Phi$ - joint feature map;\\
 $\phi_1$ - node feature map;\\
 $\phi_2$ - edge feature map;\\
 $S_n$ - constraint set for training instance $n$;\\
 $y^*$ - solution of the quadratic assignment problem;\\
 $\hat{y}$ - most violated constraint in column generation;\\ 
 $y_{ii'}$ - $i^{th}$ row and $i'^{th}$ column element of $y$;\\
 $c_{ii'}$ - value of compatibility function for map $i\mapsto i'$;\\
 $d_{ii'jj'}$ - value of compatibility function for map $ij\mapsto i'j'$;\\
 $\epsilon$ - tolerance for column generation;\\
 $w_1$ - node parameter vector; $w_2$ - edge parameter vector; $w:=[w_1~w_2]$ - joint parameter vector; $w\in \mathcal W$;\\
 $\xi_n$ - slack variable for training instance $n$;\\ 
 $\Omega$ - regularization function; $\lambda$ - regularization parameter;\\ 
 $\delta$ - convergence monitoring threshold in bistochastic normalization.
  \smallskip
  \hrule 
\end{table}

\section{The Graph Matching Problem}
\label{sec:graphmatching}

The notation used in this paper is summarized in table \ref{tab:definition}. In the following we denote a graph by $G$. We will often refer to a \emph{pair} of graphs, and the second graph in the pair will be denoted by $G'$. We study the general case of \emph{attributed} graph matching, and attributes of the
vertex $i$ and the edge $ij$ in $G$ are denoted by $G_i$ and $G_{ij}$
respectively. Standard graphs are obtained if the node
attributes are empty and the edge attributes $G_{ij}\in\{0,1\}$ are binary
denoting the absence or presence of an edge, in which case we get the so-called \emph{exact} graph matching problem.

Define a \emph{matching matrix} $y$ by $y_{ii'}\in\{0,1\}$ such that $y_{ii'}=1$ if node $i$ in the first graph maps to node $i'$ in the second graph ($i \mapsto i'$) and $y_{ii'}=0$ otherwise. Define by $c_{ii'}$ the value of the compatibility function for the unary assignment
$i \mapsto i'$ and by $d_{ii'jj'}$ the value of the compatibility function for the
pairwise assignment $ij \mapsto i'j'$. Then, a generic formulation of the graph matching problem consists of finding the optimal matching matrix $y^*$ given by the solution of the following (NP-hard) \emph{quadratic assignment problem} \cite{Anstreicher03}
\begin{align}
\label{eq:formulation}
  y^* = \argmax_{y} \left[\sum_{ii'}c_{ii'}y_{ii'} + \sum_{ii'jj'} d_{ii'jj'}y_{ii'}y_{jj'}\right], 
\end{align}
typically subject to either the injectivity constraint (one-to-one, that is $\sum_{i}
y_{ii'} \le 1$ for all $i'$, $\sum_{i'} y_{ii'} \le 1$ for all $i$) or simply the constraint that
the map should be a function (many-to-one, that is $\sum_{i'}y_{ii'}=
1$ for all $i$). If $d_{ii'jj'}=0$ for all $ii'jj'$ then \eq{eq:formulation} becomes a \emph{linear assignment problem}, exactly solvable in worst case cubic time \cite{PapSte82}. Although the compatibility functions $c$ and $d$ obviously depend on the attributes $\{G_i,G'_{i'}\}$ and $\{G_{ij},G'_{i'j'}\}$, the functional form of this dependency is typically assumed to be fixed in graph matching. This is precisely the restriction we are going to relax in this paper: both the functions $c$ and $d$ will be parametrized by vectors whose coefficients will be learned within a convex optimization framework. In a way, instead of proposing yet another algorithm for determining \emph{how} to approximate the solution for \eq{eq:formulation}, we are aiming at finding a way to determine \emph{what} should be maximized in \eq{eq:formulation}, since different $c$ and $d$ will produce different criteria to be maximized.

\section{Learning Graph Matching}
\label{sec:Setup}

\subsection{General Problem Setting}
\label{sec:Setting}

We approach the problem of learning the compatibility functions for graph matching as a supervised learning problem \cite{TsoJoaHofAlt05}. The training set comprises $N$ observations $x$ from an input set $\Xcal$, $N$ corresponding labels $y$ from an output set $\Ycal$, and can be represented by $\{(x^1;y^1),\dots,(x^N;y^N)\}$. Critical in our setting is the fact that the observations and labels are \emph{structured objects}. In typical supervised learning scenarios, observations are vectors and labels are elements from some discrete set of small cardinality, for example $y^n \in\{-1,1\}$ in the case of binary classification. However, in our case an observation $x^n$ is a \emph{pair of graphs}, i.e.~$x^n=(G^n,G'^n)$, and the label $y^n$ is a \emph{match} between graphs, represented by a matching matrix as defined in section \ref{sec:graphmatching}.

If $\Xcal=\mathcal G\times \mathcal G$ is the space of pairs of graphs, $\Ycal$ is the space of matching matrices, and $\mathcal W$ is the space of parameters of our model, then learning graph matching amounts to estimating a function $g:\mathcal G\times \mathcal G\times \mathcal W\mapsto \Ycal$ which minimizes the prediction loss on the test set. Since the test set here is assumed not to be available at training time, we use the standard approach of minimizing the empirical risk (average loss in the training set) plus a regularization term in order to avoid overfitting. The optimal predictor will then be the one which minimizes an expression of the following type:

\begin{align}
\label{eq:riskmin}
  \underbrace{\frac{1}{N}\sum_{n=1}^N\Delta(g(G^n,G'^n;w), y^n)}_{\text{empirical risk}}+\underbrace{\lambda\Omega(w)}_{\text{regularization term}},
\end{align}

\noindent where $\Delta(g(G^n,G'^n;w), y^n)$ is the loss incurred by the predictor $g$ when predicting, for training input $(G^n,G'^n)$, the output $g(G^n,G'^n;w)$ instead of the training output $y^n$. The function $\Omega(w)$ penalizes `complex' vectors $w$, and $\lambda$ is a parameter that trades off data fitting against generalization ability, which is in practice determined using a validation set. In order to completely specify such an optimization problem, we need to define the parametrized class of predictors $g(G,G';w)$ whose parameters $w$ we will optimize over, the loss function $\Delta$ and the regularization term $\Omega(w)$. In the following we will focus on setting up the optimization problem by addressing each of these points.

\subsection{The Model}
\label{subsec:model}

We start by specifying a $w$-parametrized class of predictors
$g(G,G';w)$. We use the standard approach of discriminant functions,
which consists of picking as our optimal estimate the one for which the
discriminant function $f(G,G',y;w)$ is maximal,
i.e.~$g(G,G';w)=\argmax_y f(G,G',y;w)$. We assume linear discriminant
functions $f(G,G',y;w) = \inner{w}{\Phi(G,G',y)}$, so that our predictor
has the form 
\begin{align}
\label{eq:argmax}
  g(G,G', w) = \argmax_{y \in \Ycal}\inner{w}{\Phi(G,G',y)}.
\end{align}
Effectively we are solving an
inverse optimization problem, as described by
\cite{TsoJoaHofAlt05,AhuOrl01}, that is, we are trying to 
find $f$ such that $g$ has desirable properties. 
Further specification of $g(G,G';w)$ requires determining the joint
feature map $\Phi(G,G',y)$, which has to encode the properties of both
graphs as well as the properties of a match $y$ between these
graphs. The key observation here is that we can relate the quadratic
assignment formulation of graph matching, given by \eq{eq:formulation},
with the predictor given by \eq{eq:argmax}, and interpret the solution
of the graph matching problem as being the estimate of $g$,
i.e.~$y^w=g(G,G';w)$. This allows us to interpret the discriminant
function in \eq{eq:argmax} as the objective function to be maximized in
\eq{eq:formulation}: 
\begin{align}
\label{eq:equate}
\inner{\Phi(G,G',y)}{w}=\sum_{ii'}c_{ii'}y_{ii'} + \sum_{ii'jj'} d_{ii'jj'}y_{ii'}y_{jj'}.
\end{align}
This clearly reveals that the graphs and the parameters must be encoded in
the compatibility functions. The last step before obtaining $\Phi$
consists of choosing a parametrization for the compatibility
functions. We assume a simple linear parametrization
\begin{subequations}
  \label{eq:cd3}
  \begin{align}
    \label{eq:cii}
    c_{ii'} & = \inner{\phi_1(G_i, G_{i'}')}{w_1}\\
    \label{eq:dii}
    d_{ii'jj'} & = \inner{\phi_2(G_{ij}, G_{i'j'}')}{w_2},
  \end{align}
\end{subequations}

\noindent i.e.~the compatibility functions are linearly dependent on the parameters, and on new feature maps $\phi_1$ and $\phi_2$ that only involve the graphs (section \ref{sec:Models} specifies the feature maps $\phi_1$ and $\phi_2$). As already defined, $G_i$ is the attribute of node $i$ and $G_{ij}$ is the attribute of edge $ij$ (similarly for $G'$). However, we stress here that these are not necessarily \emph{local} attributes, but are arbitrary features \emph{simply indexed} by the nodes and edges.\footnote{As a result in our general setting `node' compatibilities and `edge' compatibilities become somewhat misnomers, being more appropriately described as unary and binary compatibilities. We however stick to the standard terminology for simplicity of exposition.} For instance, we will see in section \ref{sec:Models} an example where $G_i$ encodes the graph structure of $G$ as `seen' from node $i$, or from the `perspective' of node $i$. 

Note that the traditional way in which graph matching is approached arises as a particular case of equations \eq{eq:cd3}: if $w_1$ and $w_2$ are constants, then $c_{ii'}$ and $d_{ii'jj'}$ depend only on the features of the graphs. By defining $w:=[w_1~w_2]$, we arrive at the final form for $\Phi(G,G',y)$ from \eq{eq:equate} and \eq{eq:cd3}:

\begin{multline}
\label{eq:psi}
  \Phi(G, G', y) =\\
  \Biggl[\sum_{ii'} y_{ii'} \phi_1(G_i, G_{i'}'), 
    \sum_{ii'jj'} y_{ii'} y_{jj'} \phi_2(G_{ij}, G_{i'j'}')\Biggr].
\end{multline}

\noindent Naturally, the final specification of the predictor $g$ depends on the choices of $\phi_1$ and $\phi_2$. Since our experiments are concentrated on the computer vision domain, we use typical computer vision features (e.g.~Shape Context) for constructing $\phi_1$ and a simple edge-match criterion for constructing $\phi_2$ (details follow in section \ref{sec:Models}).

\subsection{The Loss}
\label{subsec:loss}

Next we define the loss $\Delta(y,y^n)$ incurred by estimating the matching matrix $y$
 instead of the correct one, $y^n$. When both graphs have large sizes, we define this as the fraction of mismatches between matrices $y$ and $y^n$, i.e.
\begin{align}
   \label{eq:loss}
   \Delta(y,y^n) = 1 - \frac{1}{\nbr{y^n}_F^2}\sum_{ii'}y_{ii'}y^n_{ii'}.
 \end{align}
(where $\nbr{\cdot}_F$ is the Frobenius norm). If one of the graphs has a small size, this measure may be too rough. In our experiments we will encounter such a situation in the context of matching in images. In this case, we instead use the loss
\begin{align}
\Delta(G,G',\pi) =
 1 - \frac{1}{|\pi|}\sum_{i} \sbr{\frac{d(G_i, G'_{\pi(i)})} {\sigma}}.
\end{align}
Here graph nodes correspond to point sets in the images, $G$ corresponds to the smaller, `query' graph, and $G'$ is the larger, `target' graph (in this expression, $G_i$ and $G'_j$ are particular points in $G$ and $G'$; $\pi(i)$ is the index of the point in $G'$ to which the $i^{th}$ point in $G$ should be correctly mapped; $d$ is simply the Euclidean distance, and is scaled by $\sigma$, which is simply the width of the image in question). Hence we are penalising matches based on how distant they are from the correct match; this is commonly referred to as the `endpoint error'.

Finally, we specify a quadratic regularizer $\Omega(w)=\frac{1}{2}\nbr{w}^2$.

\subsection{The Optimization Problem}

Here we combine the elements discussed in \ref{subsec:model} in order to formally set up a mathematical optimization problem that corresponds to the learning procedure. The expression that arises from \eq{eq:riskmin} by incorporating the specifics discussed in \ref{subsec:model}/\ref{subsec:loss} still consists of a very difficult (in particular non-convex) optimization problem. Although the regularization term is convex in the parameters $w$, the empirical risk, i.e.~the first term in \eq{eq:riskmin}, is not. Note that there is a finite number of possible matches $y$, and therefore a finite number of possible values for the loss $\Delta$; however, the space of parameters $\mathcal W$ is continuous. What this means is that there are large equivalence classes of $w$ (an equivalence class in this case is a given set of $w$'s each of which produces the same loss). Therefore, the loss is piecewise constant on $w$, and as a result certainly not amenable to any type of smooth optimization.

One approach to render the problem of minimizing \eq{eq:riskmin} more
tractable is to replace the empirical risk by a convex upper bound on
the empirical risk, an idea that has been exploited in Machine Learning
in recent years \cite{TsoJoaHofAlt05,SmoVisQuo08,TasGueKol03}. By minimizing this convex upper
bound, we hope to decrease the empirical risk as well. It is easy to show
that the convex (in particular, linear) function $\frac{1}{N}\sum_n
\xi_n$ is an upper bound for $\frac{1}{N}\sum_n \Delta(g(G^n,G'^n;w),y^n)$
for the solution of \eq{eq:riskmin} with appropriately chosen
constraints: 
\begin{subequations}
  \label{eq:primal}
  \begin{align}
    \label{eq:primal-obj}
    \mini_{w,\xi} ~  &
    \frac{1}{N}\sum_{n=1}^N \xi_n + \frac{\lambda}{2} \nbr{w}^2 \\
    \text{subject to } 
    & \inner{w}{\Psi^n(y)} \geq \Delta(y, y^n) - \xi_n 
    \label{eq:primal-cons} \\
    \nonumber
    & \text{for all } 
    n\text{ and }  y \in \Ycal.
  \end{align}
\end{subequations}
Here we define $\Psi^n(y):=\Phi(G^n,G'^n,y^n)-\Phi(G^n,G'^n,y)$. Formally, we have:
\begin{lemma} 
For any feasible $(\xi, w)$ of \eq{eq:primal} the inequality $\xi_n \geq
\Delta(g(G^n,G'^n;w),y^n)$ holds for all $n$. In particular, for the optimal solution 
$(\xi^*,w^*)$ we have $\frac{1}{N}\sum_n
\xi^*_n \ge \frac{1}{N}\sum_n \Delta(g(G^n,G'^n;w^*),y^n)$. 
\end{lemma}
\begin{proof}
The constraint \eq{eq:primal-cons} needs to hold for all $y$, hence in
particular for $y^{w^*} = g(G^n,G'^n;w^*)$. By construction $y^{w^*}$ satisfies $\inner{w}{\Psi^n(y^{w^*})}
\leq 0$. Consequently $\xi_n \geq \Delta(y^{w^*},y^n)$. The second part of
the claim follows immediately. 
\end{proof}
%
%
%

The constraints \eq{eq:primal-cons} mean that the margin
$f(G^n,G'^n,y^n;w) - f(G^n,G'^n, y;w)$, i.e.~the gap between the discriminant functions for $y^n$ and $y$ should exceed the loss induced by estimating
$y$ instead of the training matching matrix $y^n$. This is highly intuitive since it reflects the fact that we
want to safeguard ourselves most against mis-predictions $y$ which incur
a large loss (i.e.~the smaller is the loss, the less we should care about making a mis-prediction, so we can enforce a smaller margin). The presence of $\xi_n$ in the constraints and in the objective function means that we allow the hard inequality (without $\xi_n$) to be violated, but we penalize violations for a given $n$ by adding to the objective function the cost $\frac{1}{N}\xi_n$.



Despite the fact that \eq{eq:primal} has exponentially many constraints
(every possible matching $y$ is a constraint), we will see in what
follows that there is an efficient way of finding an
$\epsilon$-approximation to the optimal solution of \eq{eq:primal} by
finding the worst violators of the constrained optimization problem.

\subsection{The Algorithm}

Note that the number of constraints in \eq{eq:primal} is given by the
number of \emph{possible} matching matrices $|\Ycal|$ times the number 
of training instances $N$. In graph matching the number of possible
matches between two graphs grows factorially with their size. In this case it
is infeasible to solve \eq{eq:primal} exactly.

There is however a way out of this problem by using an optimization
technique known as \emph{column generation} \cite{PapSte82}. Instead of solving
\eq{eq:primal} directly, one computes the most violated constraint in
\eq{eq:primal} iteratively for the current solution and
adds this constraint to the optimization problem. In order to do so,
we need to solve
\begin{align}
  \label{eq:max_violator}
  \argmax_{y} \left[\inner{w}{\Phi(G^n,G'^n,y)} + \Delta(y, y^n)\right],
\end{align}  
as this is the term for which the constraint \eq{eq:primal-cons} is
tightest (i.e.~the constraint that maximizes $\xi_n$).

The resulting algorithm is given in algorithm \ref{alg:colgen}. We use the 'Bundle Methods for Regularized Risk Minimization' (BMRM) solver of \cite{TeoLeSmoVis07}, which merely requires that for each candidate $w$, we compute the gradient of $\frac{1}{N} \sum \inner{w}{\Psi(\hat{y})} + \frac{\lambda}{2}\nbr{w}^2$ with respect to $w$, and the loss ($\frac{1}{N} \sum_n \Delta(\hat{y}, y^n)$) ($\hat{y}$ is the most violated constraint in column generation). See \cite{TeoLeSmoVis07} for further details


\begin{algorithm}[t]
  \caption{Column Generation}
  \label{alg:colgen}
\begin{algorithmic}
    \STATE {\bfseries Define:}
    \STATE $\Psi^n(y):=\Phi(G^n,G'^n,y^n)-\Phi(G^n,G'^n,y)$
    \STATE $H^n(y):=\inner{w}{\Phi(G^n,G'^n,y)}+\Delta(y,y^n)$
    \STATE {\bfseries Input:} training graph pairs $\{G^n\}$,$\{G'^n\}$, training matching matrices $\{y^n\}$,
    sample size $N$, tolerance $\epsilon$ 
    \STATE Initialize $S_n = \emptyset$ for all $n$, and $w = 0$
    \REPEAT
    \STATE Get current $w$ from BMRM
    \FOR{$n=1$ {\bfseries to} $N$} 
    \STATE $\hat{y} = \argmax_{y \in \Ycal} H^n(y)$
    \STATE Compute gradient of $\inner{w}{\Psi(G^n,G'^n,y)} + \frac{\lambda}{2}\nbr{w}^2$ w.r.t. $w$ ($= \Psi_n(\hat{y}) + \lambda w$)
    \STATE Compute loss $\Delta(\hat{y}, y^n)$
    \ENDFOR
    \STATE Report $\frac{1}{N} \sum \xi_n$ and $\frac{1}{N} \sum_n \Delta(\hat{y}, y^n)$ to BMRM
    \UNTIL{$\frac{1}{N} \sum \xi_n$ is sufficiently small}
  \end{algorithmic}
\end{algorithm}

Let us investigate the complexity of solving \eq{eq:max_violator}. Using
the joint feature map $\Phi$ as in \eq{eq:psi} and the loss as in
\eq{eq:loss}, the argument in \eq{eq:max_violator} becomes
\begin{align}
  \label{eq:max_violator_matching}
  & \inner{\Phi(G, G', y)}{w} + \Delta(y, y^n)= \\ 
  = & \sum_{ii'} y_{ii'} \bar{c}_{ii'} + \sum_{ii'jj'} y_{ii'} y_{jj'}
  d_{ii'jj'} + \mathrm{constant},
  \nonumber
\end{align}  
where $\bar{c}_{ii'} = \inner{\phi_1(G_i, G_{i'}')}{w_1} + y^n_{ii'} /
\nbr{y^n}^2_F$ and $d_{ii'jj'}$ is defined as in \eq{eq:dii}.

The maximization of \eq{eq:max_violator_matching}, which needs to be carried out at \emph{training} time, is a quadratic assignment problem, as is the problem to be solved at test time. In the particular case where $d_{ii'jj'}=0$ throughout, both the problems at training and at test time
are linear assignment problems, which can be solved efficiently in
worst case cubic time. 

In our experiments, we solve the linear assignment problem with the
efficient solver from \cite{JonVol87} (`house' sequence), and the
Hungarian algorithm (video/bikes dataset). For quadratic assignment, we
developed a C++ implementation of the well-known Graduated Assignment
algorithm \cite{Gold96}. However the learning scheme discussed here is
independent of which algorithm we use for solving either linear or
quadratic assignment. Note that the estimator is but a mere
approximation in the case of quadratic assignment: since we are unable
to find the most violated constraints of \eq{eq:max_violator}, we cannot
be sure that the duality gap is properly minimized in the constrained
optimization problem.

\section{Features for the Compatibility Functions}
\label{sec:Models}

The joint feature map $\Phi(G,G',y)$ has been derived in its full
generality \eq{eq:psi}, but in order to have a working model we need to
choose a specific form for $\phi_1(G_i,G'_{i'})$ and
$\phi_2(G_{ij},G'_{i'j'})$, as mentioned in section \ref{sec:Setup}. We
first discuss the linear features $\phi_1$ and then proceed to the quadratic terms
$\phi_2$. For concreteness, here we only discuss options actually used
in our experiments.

\subsection{Node Features}
\label{subsec:nodefeatures}
We construct $\phi_1(G_i,G'_{i'})$ using the squared difference
$\phi_1(G_i,G'_{i'})=(\dots,-|G_i(r)-G'_{i'}(r)|^2,\dots)$. This differs from what is shown in \cite{CaeCheLeSmo07}, in which an exponential decay is used (i.e.~$\exp(-|G_i(r)-G'_{i'}(r)|^2/)$); we found that using the squared difference resulted in much better performance after learning.
Here
$G_i(r)$ and $G'_{i'}(r)$ denote the $r^{th}$ coordinates of the
corresponding attribute vectors. Note that in standard graph matching
without learning we typically have $c_{ii'}=\exp(-\nbr{G_i-G'_{i'}}^2)$,
which can be seen as the particular case of \eq{eq:cii} for both
$\phi_1$ and $w_1$ flat, given by
$\phi_1(G_i,G'_{i'})=(\dots,\exp(-\nbr{G_i-G'_{i'}}^2),\dots)$ and
$w_1=(\dots,1,\dots)$ \cite{CouSriShi07}. Here instead we have
$c_{ii'}=\inner{\phi_1(G_i,G'_{i'})}{w_1}$, where $w_1$ is learned from
training data. In this way, by tuning the $r^{th}$ coordinate of $w_1$
accordingly, the learning process finds the relevance of the $r^{th}$
feature of $\phi_1$. In our experiments (to be described in the next
section), we use the well-known 60-dimensional Shape Context features
\cite{Belongie02}. They encode how each node `sees' the other
nodes. It is an instance of what we called in section \ref{sec:Setup} a
feature that captures the node `perspective' with respect to the
graph. We use 12 angular bins (for angles in $[0,\frac{\pi}{6}) \ldots [\frac{11\pi}{6},2\pi)$), and 5 radial bins (for radii in $(0,0.125), [0.125,0.25) \ldots [1,2)$, where the radius is scaled by the average of all distances in the scene) to obtain our 60 features. This is similar to the setting described in \cite{Belongie02}.

\subsection{Edge Features}

For the edge features $G_{ij}$ ($G'_{i'j'}$), we use standard graphs, i.e.~$G_{ij}$ ($G'_{i'j'}$) is $1$ if there is an edge between $i$ and $j$ and $0$ otherwise. In this case, we set $\phi_2(G_{ij},G'_{i'j'})=G_{ij}G'_{i'j'}$ (so that $w_2$ is a scalar).

\section{Experiments}
\label{sec:Experiments}

\begin{figure}[p]
\begin{center}
\includegraphics[angle=-90,width=0.8\columnwidth]{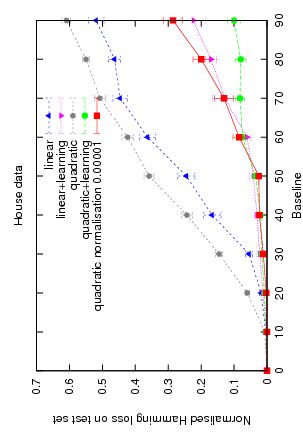}

\includegraphics[angle=-90,width=0.8\columnwidth]{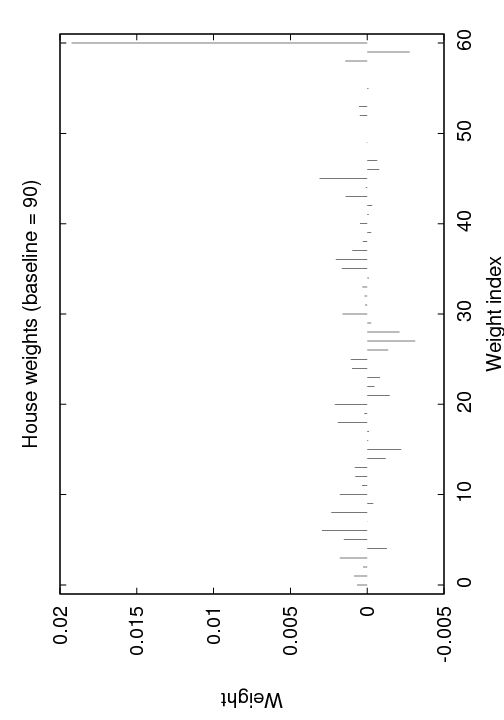}

\includegraphics[height=0.09\textheight]{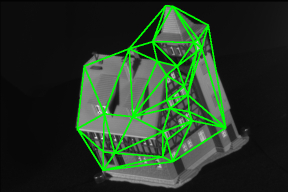}

\includegraphics[height=0.09\textheight]{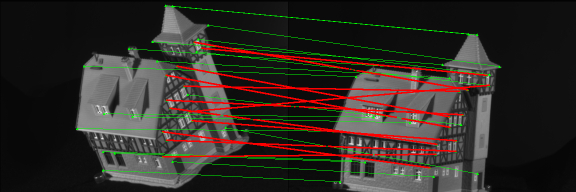}\\
\includegraphics[height=0.09\textheight]{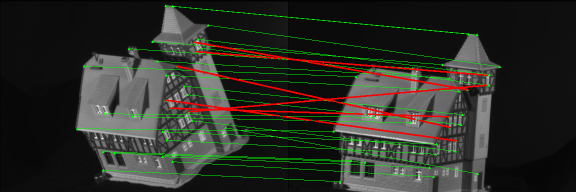}
\end{center}

\caption{
Top: Performance on the `house' sequence as the baseline (separation between frames) varies (the normalised Hamming loss on all testing examples is reported, with error bars indicating the standard error). Centre: The weights learned for the quadratic model (baseline = 90, $\lambda = 1$). Bottom: A frame from the sequence, together with its landmarks and triangulation; the 3$^{rd}$ and the 93$^{rd}$ frames, matched using linear assignment (without learning, loss = $12/30$), and the same match after learning ($\lambda = 10$, loss = $6/30$). Mismatches are shown in red.
}
\label{fig:housepics}
\end{figure}

\begin{figure}[p]
\begin{center}
\includegraphics[angle=-90,width=0.8\columnwidth]{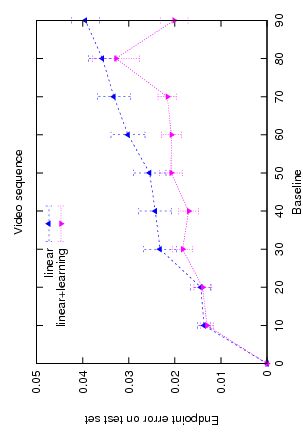}

\includegraphics[angle=-90,width=0.8\columnwidth]{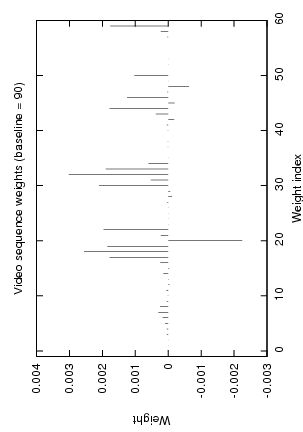}

\includegraphics[height=0.11\textheight]{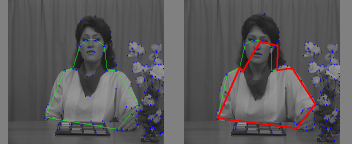}\\
\includegraphics[height=0.11\textheight]{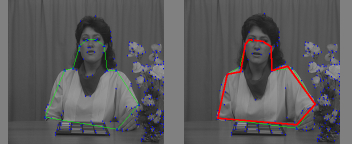}
\end{center}

\caption{
Top: Performance on the video sequence as the baseline (separation between frames) varies (the endpoint error on all testing examples is reported, with error bars indicating the standard error). Centre: The weights learned for the model (baseline = 90, $\lambda = 100$). Bottom: The 7$^{th}$ and the 97$^{th}$ frames, matched using linear assignment (loss = $0.028$), and the same match after learning ($\lambda = 100$, loss = $0.009$). The outline of the points to be matched (left), and the correct match (right) are shown in green; the inferred match is outlined in red; the match after learning is much closer to the correct match.
}
\label{fig:alexpics}
\end{figure}

\subsection{House Sequence}

For our first experiment, we consider the CMU `house' sequence -- a dataset consisting of 111 frames of a toy house \cite{CMU_house}. Each frame in this sequence has been hand-labelled, with the same 30 landmarks identified in each frame \cite{CaeCaeSchBar06}. We explore the performance of our method as the baseline (separation between frames) varies.

For each baseline (from 0 to 90, by 10), we identified all pairs of images separated by exactly this many frames. We then split these pairs into three sets, for training, validation, and testing. In order to determine the adjacency matrix for our edge features, we triangulated the set of landmarks using the Delaunay triangulation (see figure \ref{fig:housepics}).

Figure \ref{fig:housepics} (top) shows the performance of our method as the baseline increases, for both linear and quadratic assignment (for quadratic assignment we use the Graduated Assignment algorithm, as mentioned previously). The values shown report the normalised Hamming loss (i.e.~the proportion of points incorrectly matched); the regularization constant resulting in the best performance on our validation set is used for testing. Graduated assignment using bistochastic normalisation, which to the best of our knowledge is the state-of-the-art relaxation, is shown for comparison \cite{CouSriShi07}.\footnote{Exponential decay on the node features \emph{was} beneficial when using the method of \cite{CouSriShi07}, and has hence been maintained in this case (see section \ref{subsec:nodefeatures}); a normalisation constant of $\delta = 0.00001$ was used.}

For both linear and quadratic assignment, figure \ref{fig:housepics} shows that learning significantly outperforms non-learning in terms of accuracy. Interestingly, quadratic assignment performs \emph{worse} than linear assignment before learning is applied -- this is likely because the relative scale of the linear and quadratic features is badly tuned before learning. Indeed, this demonstrates exactly why learning is important. It is also worth noting that linear assignment \emph{with} learning performs similarly to quadratic assignment with bistochastic normalisation (without learning) -- this is an important result, since quadratic assignment via Graduated Assignment is significantly more computationally intensive.


Figure \ref{fig:housepics} (centre) shows the weight vector learned using quadratic assignment (for a baseline of 90 frames, with $\lambda=1$). Note that the first 60 points show the weights of the Shape Context features, whereas the final point corresponds to the edge features. The final point is given a very high score after learning, indicating that the edge features are important in this model.\footnote{This should be interpreted with some caution: the features have different scales, meaning that their importances cannot be compared directly. However, from the point of view of the regularizer, assigning this feature a high weight bares a high cost, implying that it is an important feature.} Here the first 12 features correspond to the first radial bin (as described in section \ref{sec:Models}) etc. The first radial bin appears to be more important than the last, for example. Figure \ref{fig:housepics} (bottom) also shows an example match, using the 3$^{rd}$ and the 93$^{rd}$ frames of the sequence for linear assignment, before and after learning.

\begin{figure}
 \includegraphics[angle=-90,width=\columnwidth]{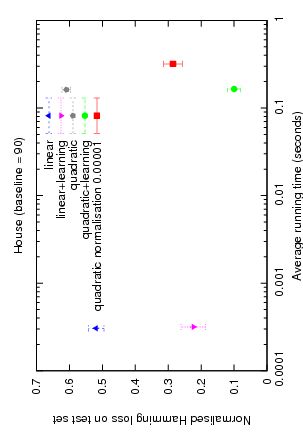}
\caption{Running time versus accuracy on the `house' dataset, for a baseline of 90. Standard errors of both running time and performance are shown (the standard error for the running time is almost zero). Note that linear assignment is around three orders of magnitude faster than quadratic assignment.}
\label{fig:times}
\end{figure}

Finally, Figure \ref{fig:times} shows the running time of our method
compared to its accuracy. Firstly, it should be noted that the use of
learning has no effect on running time; since learning outperforms
non-learning in all cases, this presents a very strong case for
learning. Quadratic assignment with bistochastic normalisation
gives the best non-learning performance, however, it is still worse than either linear or
quadratic assignment \emph{with} learning and it is significantly slower.

\subsection{Video Sequence}

For our second experiment, we consider matching features of a human in a video sequence. We used a video sequence from the SAMPL dataset \cite{SAMPL} -- a 108 frame sequence of a human face (see figure \ref{fig:alexpics}, bottom). To identify landmarks for these scenes, we used the SUSAN corner detector \cite{susan92,susan96}. This detector essentially identifies points as corners if their neighbours within a small radius are dissimilar. This detector was tuned such that no more than 200 landmarks were identified in each scene.

In this setting, we are no longer interested in matching \emph{all} of the landmarks in both images, but rather those that correspond to important parts of the human figure. We identified the same 11 points in each image (figure \ref{fig:alexpics}, bottom). It is assumed that these points are known in advance for the template scene ($G$), and are to be found in the target scene ($G'$). Clearly, since the correct match corresponds to only a tiny proportion of the scene, using the normalised Hamming loss is no longer appropriate -- we wish to penalise incorrect matches less if they are `close to' the correct match. Hence we use the loss function (as introduced in section \ref{subsec:model})
\begin{align}
\Delta(G,G',\pi) =
 1 - \frac{1}{|\pi|}\sum_{i} \sbr{\frac{d(G_i, G'_{\pi(i)})} {\sigma}}.
\end{align}
Here the loss is small if the distance between the chosen match and the correct match is small.

Since we are interested in only a few of our landmarks, triangulating the graph is no longer meaningful. Hence we present results only for linear assignment.

Figure \ref{fig:alexpics} (top) shows the performance of our method as the baseline increases. In this case, the performance is non-monotonic as the subject moves in and out of view throughout the sequence. This sequence presents additional difficulties over the `house' dataset, as we are subject to noise in the detected landmarks, and possibly in their labelling also. Nevertheless, learning outperforms non-learning for all baselines. The weight vector (figure \ref{fig:alexpics}, centre) is heavily peaked about particular angular bins.

\subsection{Bikes}

\begin{figure}
  \parbox{\columnwidth}
  {
   \begin{center}
   \includegraphics[width=0.25\columnwidth]{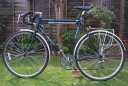}\includegraphics[width=0.25\columnwidth]{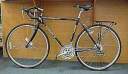}\includegraphics[width=0.25\columnwidth]{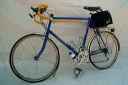}\includegraphics[width=0.25\columnwidth]{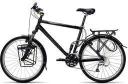}\\
   \includegraphics[width=0.25\columnwidth]{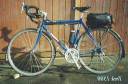}\includegraphics[width=0.25\columnwidth]{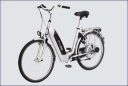}\includegraphics[width=0.25\columnwidth]{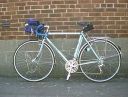}\includegraphics[width=0.25\columnwidth]{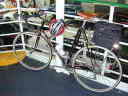}\\
   \includegraphics[width=0.25\columnwidth]{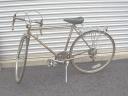}\includegraphics[width=0.25\columnwidth]{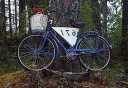}\includegraphics[width=0.25\columnwidth]{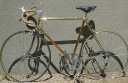}\includegraphics[width=0.25\columnwidth]{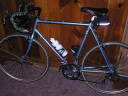}\\
   \includegraphics[width=0.25\columnwidth]{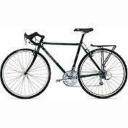}\includegraphics[width=0.25\columnwidth]{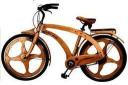}\includegraphics[width=0.25\columnwidth]{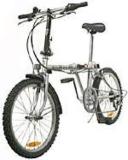}\includegraphics[width=0.25\columnwidth]{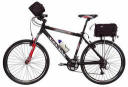}
    \end{center}
  }
  \parbox{\columnwidth}
  {
   \begin{center}
   \includegraphics[width=\columnwidth]{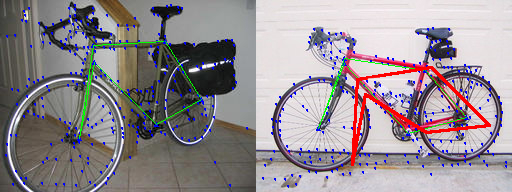}\\
   \includegraphics[width=\columnwidth]{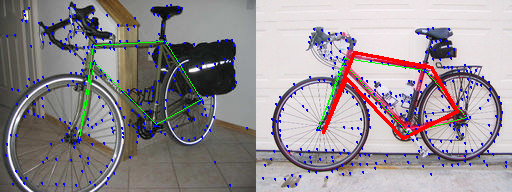}
   \end{center}
  }
\caption{Top: Some of our training scenes. Bottom: A match from our test set. The top frame shows the points as matched without learning (loss = $0.105$), and the bottom frame shows the match with learning (loss = $0.038$). The outline of the points to be matched (left), and the correct match (right) are outlined in green; the inferred match is outlined in red.}
\label{fig:bikepics}
\end{figure}

For our final experiment, we used images from the Caltech 256 dataset \cite{griffinHolubPerona}. We chose to match images in the `touring bike' class, which contains 110 images of bicycles. Since the Shape Context features we are using are robust to only a small amount of rotation (and not to reflection), we only included images in this dataset that were taken `side-on'. Some of these were then reflected to ensure that each image had a consistent orientation (in total, 78 images remained). Again, the SUSAN corner detector was used to identify the landmarks in each scene; 6 points corresponding to the frame of the bicycle were identified in each frame (see figure \ref{fig:bikepics}, bottom).

Rather than matching all \emph{pairs} of bicycles, we used a fixed template ($G$), and only varied the target. This is an easier problem than matching all pairs, but is realistic in many scenarios, such as image retrieval.

Table \ref{table:bikeresults} shows the endpoint error of our method, and gives further evidence of the improvement of learning over non-learning. Figure \ref{fig:bikepics} shows a selection of data from our training set, as well as an example matching, with and without learning.

\begin{table}
\centering

\begin{tabular}{|l|l|l|}
\hline
           & Loss & Loss (learning)\\
\hline
Training   & 0.094 (0.005) & 0.057 (0.004)\\
\hline
Validation & 0.040 (0.007) & 0.040 (0.006)\\
\hline
Testing    & 0.101 (0.005) & \textbf{0.062 (0.004)}\\
\hline
\end{tabular}
\caption{Performance on the `bikes' dataset. Results for the minimiser of the validation loss ($\lambda = 10000$) are reported. Standard errors are in parentheses.}
\label{table:bikeresults}
\end{table}

\section{Conclusions and Discussion}
\label{sec:Discussion}

We have shown how the compatibility functions for the graph matching problem can be estimated from labeled training examples, where a training input is a pair of graphs and a training output is a matching matrix. We use large-margin structured estimation techniques with column generation in order to solve the learning problem efficiently, despite the huge number of constraints in the optimization problem. We presented experimental results in three different settings, each of which revealed that the graph matching problem can be significantly improved by means of learning. 

An interesting finding in this work has been that \emph{linear} assignment with learning performs similarly to Graduated Assignment with bistochastic normalisation, a state-of-the-art \emph{quadratic} assignment relaxation algorithm. This suggests that, in situations where speed is a major issue, linear assignment may be resurrected as a means for graph matching. In addition to that, if learning is introduced to Graduated Assignment \emph{itself}, then the performance of graph matching \emph{improves significantly} both on accuracy and speed when compared to the best existing quadratic assignment relaxation \cite{CouSriShi07}.






There are many other situations in which learning a matching criterion
can be useful. In multi-camera settings for example, when different
cameras may be of different types and have different calibrations and
viewpoints, it is reasonable to expect that the optimal compatibility
functions will be different depending on which camera pair we
consider. In surveillance applications we should take advantage of the
fact that much of the context does not change: the camera and the
viewpoint are typically the same.  

To summarize, by learning a matching criterion from previously labeled
data, we are able to substantially improve the accuracy of graph matching algorithms.





\begin{thebibliography}{10}
\providecommand{\url}[1]{#1}
\csname url@samestyle\endcsname
\providecommand{\newblock}{\relax}
\providecommand{\bibinfo}[2]{#2}
\providecommand{\BIBentrySTDinterwordspacing}{\spaceskip=0pt\relax}
\providecommand{\BIBentryALTinterwordstretchfactor}{4}
\providecommand{\BIBentryALTinterwordspacing}{\spaceskip=\fontdimen2\font plus
\BIBentryALTinterwordstretchfactor\fontdimen3\font minus
  \fontdimen4\font\relax}
\providecommand{\BIBforeignlanguage}[2]{{%
\expandafter\ifx\csname l@#1\endcsname\relax
\typeout{** WARNING: IEEEtran.bst: No hyphenation pattern has been}%
\typeout{** loaded for the language `#1'. Using the pattern for}%
\typeout{** the default language instead.}%
\else
\language=\csname l@#1\endcsname
\fi
#2}}
\providecommand{\BIBdecl}{\relax}
\BIBdecl

\bibitem{CaeCheLeSmo07}
T.~S. Caetano, L.~Cheng, Q.~V. Le, and A.~J. Smola, ``Learning graph
  matching,'' in \emph{International Conference on Computer Vision}, 2007.

\bibitem{LeoHeb05}
M.~Leordeanu and M.~Hebert, ``A spectral technique for correspondence problems
  using pairwise constraints,'' in \emph{ICCV}, 2005.

\bibitem{Wang04}
H.~Wang and E.~R. Hancock, ``A kernel view of spectral point pattern
  matching,'' in \emph{International Workshops SSPR \& SPR, LNCS 3138}, 2004,
  pp. 361--369.

\bibitem{Shapiro92}
L.~Shapiro and J.~Brady, ``Feature-based correspondence - an eigenvector
  approach,'' \emph{Image and Vision Computing}, vol.~10, pp. 283--288, 1992.

\bibitem{Carcassoni03}
M.~Carcassoni and E.~R. Hancock, ``Spectral correspondence for point pattern
  matching,'' \emph{Pattern Recognition}, vol.~36, pp. 193--204, 2003.

\bibitem{Caelli04}
T.~Caelli and S.~Kosinov, ``An eigenspace projection clustering method for
  inexact graph matching,'' \emph{IEEE Trans. PAMI}, vol.~26, no.~4, pp.
  515--519, 2004.

\bibitem{Caetano04Graphical}
T.~S. Caetano, T.~Caelli, and D.~A.~C. Barone, ``Graphical models for graph
  matching,'' in \emph{IEEE International Conference on Computer Vision and
  Pattern Recognition}, Washington, DC, 2004, pp. 466--473.

\bibitem{Rosenfeld82}
A.~Rosenfeld and A.~C. Kak, \emph{Digital Picture Processing}.\hskip 1em plus
  0.5em minus 0.4em\relax New York, NY: Academic Press, 1982.

\bibitem{Wilson97}
R.~C. Wilson and E.~R. Hancock, ``Structural matching by discrete relaxation,''
  \emph{IEEE Trans. PAMI}, vol.~19, no.~6, pp. 634--648, 1997.

\bibitem{Hancock02}
E.~Hancock and R.~C. Wilson, ``Graph-based methods for vision: A yorkist
  manifesto,'' \emph{SSPR \& SPR 2002, LNCS}, vol. 2396, pp. 31--46, 2002.

\bibitem{Christmas95}
W.~J. Christmas, J.~Kittler, and M.~Petrou, ``Structural matching in computer
  vision using probabilistic relaxation,'' \emph{IEEE Trans. PAMI}, vol.~17,
  no.~8, pp. 749--764, 1994.

\bibitem{Kittler89}
J.~V. Kittler and E.~R. Hancock, ``Combining evidence in probabilistic
  relaxation,'' \emph{Int. Journal of Pattern Recognition and Artificial
  Intelligence}, vol.~3, pp. 29--51, 1989.

\bibitem{Li94}
S.~Z. Li, ``A markov random field model for object matching under contextual
  constraints,'' in \emph{International Conference on Computer Vision and
  Pattern Recognition}, 1994, pp. 866--869.

\bibitem{Schellewald04}
C.~Schellewald, ``Convex mathematical programs for relational matching of
  object views,'' Ph.D. dissertation, University of Mannhein, 2004.

\bibitem{Pelillo99a}
M.~Pelillo, ``Replicator equations, maximal cliques, and graph isomorphism,''
  \emph{Neural Comput.}, vol.~11, pp. 1933--1955, 1999.

\bibitem{Messmer98}
B.~T. Messmer and H.~Bunke, ``A new algorithm for error-tolerant subgraph
  isomorphism detection,'' \emph{IEEE Trans. PAMI}, vol.~20, no.~5, pp.
  493--503, 1998.

\bibitem{Gold96}
S.~Gold and A.~Rangarajan, ``A graduated assignment algorithm for graph
  matching,'' \emph{IEEE Trans. PAMI}, vol.~18, no.~4, pp. 377--388, 1996.

\bibitem{Rangarajan99}
A.~Rangarajan, A.~Yuille, and E.~Mjolsness, ``Convergence properties of the
  softassign quadratic assignment algorithm,'' \emph{Neural Computation},
  vol.~11, pp. 1455--1474, 1999.

\bibitem{Sinkhorn64}
R.~Sinkhorn, ``A relationship between arbitrary positive matrices and doubly
  stochastic matrices,'' \emph{Ann. Math. Statis.}, vol.~35, pp. 876--879,
  1964.

\bibitem{Pelillo94}
M.~Pelillo and M.~Refice, ``Learning compatibility coefficients for relaxation
  labeling processes,'' \emph{IEEE Trans. on PAMI}, vol.~16, no.~9, pp.
  933--945, 1994.

\bibitem{LacTasKleJor06}
S.~Lacoste-Julien, B.~Taskar, D.~Klein, and M.~Jordan, ``Word alignment via
  quadratic assignment,'' in \emph{HLT-NAACL06}, 2006.

\bibitem{CouSriShi07}
T.~Cour, P.~Srinivasan, and J.~Shi, ``Balanced graph matching,'' in
  \emph{NIPS}, 2006.

\bibitem{Anstreicher03}
K.~Anstreicher, ``Recent advances in the solution of quadratic assignment
  problems,'' \emph{Math. Program., Ser. B}, vol.~97, pp. 27--42, 2003.

\bibitem{PapSte82}
C.~Papadimitriou and K.~Steiglitz, \emph{Combinatorial Optimization :
  Algorithms and Complexity}.\hskip 1em plus 0.5em minus 0.4em\relax {Dover
  Publications}, July 1998.

\bibitem{TsoJoaHofAlt05}
I.~Tsochantaridis, T.~Joachims, T.~Hofmann, and Y.~Altun, ``Large margin
  methods for structured and interdependent output variables,'' \emph{J. Mach.
  Learn. Res.}, vol.~6, pp. 1453--1484, 2005.

\bibitem{AhuOrl01}
R.~K. Ahuja and J.~B. Orlin, ``Inverse optimization,'' \emph{Operations
  Research}, vol.~49, no.~5, pp. 771--783, 2001.

\bibitem{SmoVisQuo08}
A.~Smola, S.~V.~N. Vishwanathan, and Q.~Le, ``Bundle methods for machine
  learning,'' in \emph{Advances in Neural Information Processing Systems 20},
  J.~Platt, D.~Koller, Y.~Singer, and S.~Roweis, Eds.\hskip 1em plus 0.5em
  minus 0.4em\relax Cambridge, MA: MIT Press, 2008, pp. 1377--1384.

\bibitem{TasGueKol03}
B.~Taskar, C.~Guestrin, and D.~Koller, ``Max-margin markov networks,'' in
  \emph{Advances in Neural Information Processing Systems 16}, S.~Thrun,
  L.~Saul, and B.~{Sch\"{o}lkopf}, Eds.\hskip 1em plus 0.5em minus 0.4em\relax
  Cambridge, MA: MIT Press, 2004.

\bibitem{TeoLeSmoVis07}
C.~Teo, Q.~Le, A.~Smola, and S.~Vishwanathan, ``A scalable modular convex
  solver for regularized risk minimization,'' in \emph{Knowledge Discovery and
  Data Mining KDD'07}, 2007.

\bibitem{JonVol87}
R.~Jonker and A.~Volgenant, ``A shortest augmenting path algorithm for dense
  and sparse linear assignment problems,'' \emph{Computing}, vol.~38, no.~4,
  pp. 325--340, 1987.

\bibitem{Belongie02}
S.~Belongie, J.~Malik, and J.~Puzicha, ``Shape matching and object recognition
  using shape contexts,'' \emph{IEEE Trans. on PAMI}, vol.~24, no.~24, pp.
  509--521, 2002.

\bibitem{CMU_house}
CMU `house' dataset:\\vasc.ri.cmu.edu/idb/html/motion/house/index.html.

\bibitem{CaeCaeSchBar06}
T.~S. Caetano, T.~Caelli, D.~Schuurmans, and D.~A.~C. Barone, ``Graphical
  models and point pattern matching,'' \emph{IEEE Trans. on PAMI}, vol.~28,
  no.~10, pp. 1646--1663, 2006.

\bibitem{SAMPL}
SAMPLE motion dataset:\\http://sampl.ece.ohio-state.edu/database.htm.

\bibitem{susan92}
S.~Smith, ``A new class of corner finder,'' in \emph{BMVC}, 1992, pp. 139--148.

\bibitem{susan96}
------, ``Flexible filter neighbourhood designation,'' in \emph{ICPR}, 1996,
  pp. 206--212.

\bibitem{griffinHolubPerona}
\BIBentryALTinterwordspacing
G.~Griffin, A.~Holub, and P.~Perona, ``Caltech-256 object category dataset,''
  California Institute of Technology, Tech. Rep. 7694, 2007. [Online].
  Available: \url{http://authors.library.caltech.edu/7694}
\BIBentrySTDinterwordspacing

\end{thebibliography}
\end{document}